\documentclass[letterpaper, 10 pt, conference]{ieeeconf}  

\IEEEoverridecommandlockouts                              

\usepackage{graphicx} 
\usepackage{amsmath} 
\usepackage{amssymb}  
\usepackage{siunitx}
\usepackage{todonotes}
\usepackage{placeins}
\usepackage{ifthen}
\usepackage{makecell}

\usepackage[caption=false,font=footnotesize]{subfig}
\usepackage{booktabs}
\usepackage{hyperref}

\newboolean{PrintVersion}
\setboolean{PrintVersion}{true}
\newcommand{\ReviewChangeRemoved}[1]{\sout{#1}}
\newcommand{\ReviewChange}[1]{\textcolor{blue}{#1}}
\ifthenelse{\boolean{PrintVersion}}{
	\renewcommand{\ReviewChangeRemoved}[1]{}
	\renewcommand{\ReviewChange}[1]{#1}
}{}

\title{\LARGE \bf
Extrinsic Camera Calibration with Semantic Segmentation
}

\author{Alexander Tsaregorodtsev$^{1}$, Johannes~Müller$^{1}$,	Jan~Strohbeck$^{1}$, Martin~Herrmann$^{1}$, \\ Michael~Buchholz$^{1}$ and Vasileios Belagiannis$^{2}$
\thanks{*Part of this work was financially supported by the Federal Ministry of Economic Affairs and Energy of Germany within the program "New Vehicle and System Technologies" (project LUKAS, grant number 19A20004F). Part of this work has been conducted as part of ICT4CART project which has received funding from the European Union’s Horizon 2020 research \& innovation program under grant agreement No. 768953. Content reflects only the authors’ view and European Commission is not responsible for any use that may be made of the information it contains. }
\thanks{$^{1}$The authors are with the Institute of Measurement, Control and Microtechnology,
        Ulm University, D-89081 Ulm, Germany \newline
        \{$<$first\,\,name$>$.$<$family\,\,name$>$@uni-ulm.de\}
}
\thanks{$^{2}$Vasileios Belagiannis is with the Department of Simulation and Graphics,
        University Magdeburg, D-39106 Magdeburg, Germany \newline
        \{vasileios.belagiannis@ovgu.de\}. Most of this work was done during his time at Ulm University.
}}%

\begin{document}

\newcommand\copyrighttextinitial{%

	\scriptsize This work has been submitted to the IEEE for possible publication. Copyright may be transferred without notice, after which this version may no longer be accessible.}%
\newcommand\copyrighttextfinal{%
	
	\scriptsize\copyright\ 2022 IEEE. Personal use of this material is permitted. Permission from IEEE must be obtained for all other uses, in any current or future media, including reprinting/republishing this material for advertising or promotional purposes, creating new collective works, for resale or redistribution to servers or lists, or reuse of any copyrighted component of this work in other works.}
\newcommand\copyrightnotice{%

	\begin{tikzpicture}[remember picture,overlay]%

	\node[anchor=south,yshift=10pt] at (current page.south) {{\parbox{\dimexpr\textwidth-\fboxsep-\fboxrule\relax}{\copyrighttextfinal}}};%
	\end{tikzpicture}%


}

\maketitle
\copyrightnotice%
\thispagestyle{empty}
\pagestyle{empty}

\begin{abstract}

Monocular camera sensors are vital to intelligent vehicle operation and automated driving assistance and are also heavily employed in traffic control infrastructure. Calibrating the monocular camera, though, is time-consuming and often requires significant manual intervention. In this work, we present an extrinsic camera calibration approach that automatizes the parameter estimation by utilizing semantic segmentation information from images and point clouds. Our approach relies on a coarse initial measurement of the camera pose and builds on lidar sensors mounted on a vehicle
with high-precision
localization to capture a point cloud of the camera environment. Afterward, a mapping between the camera and world coordinate spaces is obtained by performing a lidar-to-camera registration of the semantically segmented sensor data. We evaluate our method on simulated and real-world data to demonstrate low error measurements in the calibration results. Our approach is suitable for infrastructure sensors as well as vehicle sensors, while it does not require motion of the camera platform.

\end{abstract}

\section{INTRODUCTION}

Monocular cameras are a keystone to cost-efficient sensor setups \cite{rinner2008introduction} for both intelligent vehicles (IVs) and smart infrastructure, supporting automated driving. The calibration of such cameras is vital to their operation out in the field and, therefore, of utmost importance for applications such as self-localization~\cite{engel2019deeplocalization} or recognition~\cite{wiederer2020traffic}. However, while the intrinsic camera calibration can easily be accomplished once in a constrained environment, the recovery of the extrinsic calibration in the field or between vehicle sensors is still a challenging process. In particular, classical calibration methods \cite{andrew2001multiple} often require manually placed calibration targets in the scene. These approaches, therefore, cannot easily be applied to traffic cameras, observing densely populated junctions or to IVs out in the field. Thus, an automatic calibration solution, which works for single infrastructure cameras, is needed to reduce calibration effort. Such a solution can be used to calibrate cameras mounted on IVs as well.

Recently, sensor-to-sensor registration methods, e.g. \cite{datondji2016rotation, zhu2016robust}, have been proposed to automate the calibration process. However, these methods often require broadly overlapping fields of view (FOVs), which implies redundant sensors and hence negates the cost savings achieved by a single sensor. For IVs, specific fully automated calibration schemes, such as \cite{horn2021online}, have been developed, which rely on the motion of the sensor platform, i.e., the IV, to optimize the calibration during each time step.
\begin{figure}
    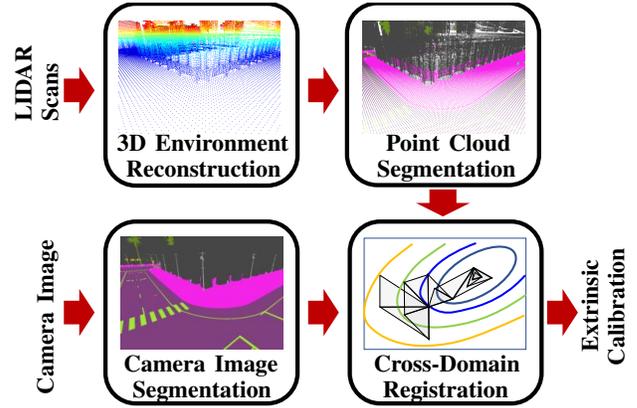
    \caption{Approach Overview. On the left side, data from both domains are fed into the pipeline. For lidar data, the 3D environment is reconstructed using multiple scans. Afterward, both domains are labeled semantically. In the final step, our proposed optimization is performed to register both domains to retrieve the extrinsic calibration parameters.}
    \label{fig:PipelineOverview}
\end{figure}
~A caveat of these specific calibration schemes is their dependency on sensor motion, which excludes static infrastructure from the range of possible applications. Deep learning approaches like \cite{li2021deepi2p} have recently been developed to estimate extrinsic camera parameters from camera images and lidar data. While these approaches have shown promising results, they still tend to have higher estimation errors compared to previous methods like \cite{horn2021online}, making them unsuitable for IV and infrastructure calibration. Finally, another classical method is described in \cite{ataer2014calibration}. It localizes a camera sensor inside of a 3D model of the environment taken from a SLAM with stereo cameras.
However, this method obtains its 3D environments from a mobile stereo camera setup and thus contains texture information not present in lidar data.
~Furthermore, this approach is not transferable to stereo camera setups permanently mounted on IVs due to their limited field of view.

We propose a monocular camera calibration approach to perform cross-domain registration of a semantically segmented mono camera image and a semantically labeled 3D model of the environment the camera is located in. An overview of the approach is given in Fig. \ref{fig:PipelineOverview}.
By using a semantic representation of both the camera and the point cloud domain, we introduce semantic labels, which can be matched between the domains. Furthermore, the use of semantic labels instead of raw RGB and RGBD camera data makes the registration more robust and reduces the possibility of registration algorithm errors. These can occur with raw camera data due to color space and exposure differences between different camera types.

Our approach consists of data pre-processing steps as well as an optimization step performing cross-domain registration. In the pre-processing steps, the 3D environment model is reconstructed and then labeled semantically using neural networks. The segmentation is also performed on images of the target camera that we aim to calibrate. Then, an optimization pass is invoked to match the visual appearance and projection of a rendered view of the 3D model with the segmented camera view.
This optimization then yields the extrinsic camera calibration.
~Finally, experiments with the CARLA simulator \cite{Dosovitskiy17} and the KITTI dataset \cite{geiger2013vision} show that our method leads to the accuracy level that is suitable for automated driving applications, both for static and dynamic camera platforms. The contribution of this paper is twofold. From a practical perspective, our approach allows for low-effort, highly automated calibration of monocular cameras in (geo-referenced) world coordinates without requiring motion of the target sensor platform.
From a methodological perspective to the best of our knowledge, we are the first to use semantic segmentation as a visual feature set to guide the required parameter optimization. The source code of our approach is publicly available\footnote{\url{https://github.com/Tuxacker/semantic\_calibration}}.

The related work to extrinsic camera calibration is outlined in Section II. The complete algorithm is described in Section III, while in Section IV, a theoretical and practical evaluation is provided. At last, the conclusion of this work is described in Section V.
\section{RELATED WORK}

Calibration of cameras has been extensively investigated in the past. A classical approach for intrinsic and extrinsic camera calibration is presented in \cite{andrew2001multiple}. One of the first approaches specifically for infrastructure sensors was described in \cite{datondji2016rotation}, where a vehicle target was used to calibrate a pair of fisheye cameras with a large overlapping field of view. Another approach using planar structure from motion and textured environments was presented in \cite{zhu2016robust}, which is able to calibrate multi-sensor configurations. The approach of \cite{ataer2014calibration} circumvents the field of view (FoV) overlap requirement of the other two approaches by reconstructing a 3D environment using a mobile RGB-D camera setup. Then, a matching of feature descriptors between the target camera image and the model is performed and the closest matching keyframes are found to determine the camera pose. The automatic calibration method of \cite{dubska2014fully} estimates the camera vanishing points by tracking image features of moving vehicles, while a newer method by \cite{bartl2021automatic} uses a neural network to find landmark positions on vehicles. By using the positional relations between landmarks obtained from real 3D models of the vehicles, a calibration is estimated from the landmarks.

Another class of algorithms performs direct 2D to 3D matching and registration. Feng et al. \cite{feng20192d3d} use a neural network to directly match point cloud patches to image patches, while the algorithm in \cite{yu2020monocular} performs camera localization by finding correspondences between 2D and 3D line features. Li et al. \cite{li2021deepi2p} train a classifier to detect if a point in the point cloud is in the camera frustum and then solve the inverse camera projection problem using the frustum labels.

Methods performing calibration through sensor motion can also be found in the literature (\cite{horn2021online, wei2018calibration, wise2020certifiably}). In particular, Horn et al. \cite{horn2021online} use a visual SLAM approach to estimate the camera displacement and then represent the translation and rotation using dual quaternions instead of homogeneous transformation matrices, which allows for a more efficient implementation of the optimization. The optimization consists of finding a calibration matrix that is able to close the loop between the current and previous time step, where the calibration matrices of the previous time step are known.\\
A recent method using automated vehicles as calibration targets is shown in \cite{muller2019laci}. It uses lidar measurements in an empty scene to determine the ground plane and background and then regresses the position of the automated vehicle as it passes through the sensors FoV with localization data of the vehicle being available.

We improve on previous methods by exploring a new concept for matching cross-domain features by visual alignment, which does not require specific calibration targets or motion of the sensor itself. It, therefore, allows us to perform a single-shot calibration in diverse scenarios without requiring a lot of manual intervention.

\section{CALIBRATION METHOD}

In this section, we formulate our problem and present each step of our approach required to estimate the extrinsic camera calibration.

\subsection{Problem Formulation and Constraints}

Consider the perspective camera model \cite{andrew2001multiple}
\begin{equation}
	{\begin{bmatrix}
	u \\ v \\ 1
	\end{bmatrix}} =
	\underbrace{\boldsymbol{K}_\text{int} \;
	\begin{bmatrix}
	\boldsymbol{R} & \boldsymbol{t} \\
	\boldsymbol{0} & 1
	\end{bmatrix}}_{:= \boldsymbol{P}}
	{\begin{bmatrix}
	x \\ y \\ z \\ 1
	\end{bmatrix}} \; ,
\end{equation}
where $[u, v, 1]^T$ denotes a homogeneous camera image coordinate of an image $\mathcal{I}$ with pixel color value $\mathcal{I}_{u,v} \in [0;1]^3$, and $[x, v, z, 1]^T\in\mathcal{M}^{3\text{D}}$ is a homogeneous point coordinate of a 3D model $\mathcal{M}^{3\text{D}}$ of the observed scene. $\mathcal{M}^{3\text{D}}$ is represented as a point cloud. Furthermore, it is assumed that the scene visible in the camera image is overlapping with the scene represented in the point cloud.

Our goal is to recover the parameters of the extrinsic camera matrix $\boldsymbol{P}$, i.e., the rotation matrix $\boldsymbol{R} \in \mathbb{R}^{3 \times 3}$ and the translation vector $\boldsymbol{t} \in \mathbb{R}^{3}$. At the same time we assume that the intrinsic parameters $\boldsymbol{K}_\text{int} \in  \mathbb{R}^{3 \times 4}$, which represent the mapping from 3D coordinates in the camera coordinate frame to 2D pixel coordinates, are already known, e.g.~as described in~\cite{andrew2001multiple}. Below, we propose a new calibration approach to estimate these extrinsic camera parameters. 

\subsection{Approach Overview}

We use visual matching of semantic information for the parameter estimation of the camera rotation $\boldsymbol{R}$ and translation $\boldsymbol{t}$.
Our method consists of four steps, starting with the 3D environment reconstruction $\mathcal{M}^{3\text{D}}$, represented as a point cloud (see Sec.~\ref{lbl:3drec}). It is recorded with the lidar sensor mounted on a vehicle. Next, the point cloud segmentation step (presented in Sec.~\ref{lbl:pclseg}) follows to extract semantic information for each 3D point. Then, the camera image segmentation (see Sec.~\ref{lbl:camseg}) assigns a class category to each pixel. Given the segmented 3D point cloud and image, we propose a cross-domain registration approach (explained in Sec.~\ref{lbl:cdreg}) to estimate the extrinsic camera parameters by visually matching the camera segmentation with a rendered point cloud view using linear programming~\cite{nelder1965simplex}.

\subsection{3D Environment Reconstruction}
\label{lbl:3drec}

We assume access to a vehicle equipped with a lidar sensor and high-precision localization to construct the 3D environment model $\mathcal{M}^{3\text{D}}$. Several lidar scans are recorded as point clouds and expressed in the world coordinate frame. The transformation from lidar sensor coordinates to world coordinates requires known lidar extrinsics.
 Due to road bumps and inaccurate orientation measurements, points in the sensor's far field come with increased positional errors, resulting in noisy 3D point clouds with many outliers. Thus, we filter out points that exceed the maximum distance $d_\text{max}$ from the vehicle's lidar sensor. In practice, $d_\text{max}$ is chosen to be between $50~\text{m}$ and $75~\text{m}$, which are standard cutoff values for Velodyne sensors like the VLP-32C or the HDL-64E used in the KITTI \cite{geiger2013vision} dataset. Next, we rely on the multi-way Iterative Closest Point (ICP) registration algorithm~\cite{chen1992object, choi2015robust} between all scans, which partially compensates for localization errors that accumulate over time. To accelerate the registration, instead of registering every scan against every other scan, all scans are sorted by their capture time, and groups of three neighboring scans are recursively registered and merged. The recursion ends when all groups are combined into the single point cloud $\mathcal{M}^{3\text{D}}$. At last, the point cloud is down-sampled during each recursion step in order to relax the ICP computation requirements. The reconstruction step can be skipped if the 3D environment is obtained by another mapping algorithm or from a different source.

\subsection{Point Cloud Segmentation}
\label{lbl:pclseg}
In the next step, semantic labels are assigned to the points of $\mathcal{M}^{3\text{D}}$ to obtain the semantic segmentation map $\mathcal{M}^{3\text{D}}_{\boldsymbol{c}}$, which contains a color label for each point in $\mathcal{M}^{3\text{D}}$. For this task, we rely on a pre-trained deep neural network to perform semantic segmentation, namely Cylinder3D~\cite{zhu2021cylindrical}, \ReviewChange{which supports segmentation of classes defined in SemanticKITTI~\cite{behley2019semantickitti}}.
\\Furthermore, the segmentation is also used to filter out dynamic object classes like cars and pedestrians.
 This leaves the buildings, ground points, vegetation, fences, poles, and traffic signs classes for registration.
Therefore, the calibration method does not require exact time synchronicity between the lidar scans and the camera image, as points belonging to dynamic objects can easily be filtered out by their label.

\subsection{Camera Image Segmentation}
\label{lbl:camseg}
After labeling the point cloud $\mathcal{M}^{3\text{D}}$, the semantic segmentation map $\mathcal{I}^\text{sem}$ is extracted from the camera image $\mathcal{I}$. Similar to the point cloud segmentation, we rely on a pre-trained deep neural network to extract semantic labels \ReviewChange{available in the Cityscapes dataset~\cite{cordts2016cityscapes}}, e.g.,~OCRNet~\cite{yuan2021segmentation}. \ReviewChange{We chose to use Cityscapes labels, as they share most classes available in SemanticKITTI and thus enable direct matching of class labels between domains.} Again, dynamic object class categories, e.g.,~cars or pedestrians, are removed from the segmentation map. For the remaining object categories, we perform a category alignment between the point cloud~\cite{zhu2021cylindrical} and the image~\cite{yuan2021segmentation} segmentation models by neglecting points and pixels with labels that are unique to this domain. It should be noted that the removal of dynamic objects results in holes in the resulting map, especially in crowded scenes. To minimize their impact, we introduce the normalization factor $\beta$ during the registration step in Section~\ref{lbl:cdreg}.

The image segmentation can also be done semi-automatically to improve calibration quality. In this case, the resulting segmentation is manually edited to fix incorrectly labeled regions in order to improve data quality and hence reduce the overall calibration error.  

\subsection{Cross-domain registration}
\label{lbl:cdreg}
Next, we aim to estimate the rotation matrix $\boldsymbol{R}$ and translation vector $\boldsymbol{t}$. We formulate this estimation problem as an optimization, where a loss correlated to the error $\lVert\hat{\boldsymbol{R}} - \boldsymbol{R}\rVert$ and $\lVert\hat{\boldsymbol{t}} - \boldsymbol{t}\rVert$ is minimized using the simplex algorithm by Nelder and Mead \cite{nelder1965simplex}.
As the Nelder-Mead method does not require gradient information, it allows a broad range of data transforms to be applied to $\mathcal{M}^{3\text{D}}$. This includes rendering transforms, which can be used to obtain a rasterized image $\hat{\mathcal{I}}$ of a specific perspective view of $\mathcal{M}^{3\text{D}}$. We, therefore, define the rendering function $\boldsymbol{f}$ as
\begin{equation}
	\hat{\mathcal{I}} = \boldsymbol{f} \big(\mathcal{M}^{3\text{D}},\boldsymbol{P}(\hat{\boldsymbol{R}},\hat{\boldsymbol{t}}), \mathcal{M}^{3\text{D}}_{\boldsymbol{c}} \big) \; ,
\end{equation}
where $\mathcal{M}^{3\text{D}}_{\boldsymbol{c}}$ is the color encoding of the segmentation and $\hat{\boldsymbol{R}}$, $\hat{\boldsymbol{t}}$ are estimates of $\boldsymbol{R}$ and $\boldsymbol{t}$. This rendering function is used to transform the point cloud $\mathcal{M}^{3\text{D}}$ with color labels $\mathcal{M}^{3\text{D}}_{\boldsymbol{c}}$ into an image $\hat{\mathcal{I}}$ with the viewpoint defined by the perspective camera matrix $\boldsymbol{P}(\hat{\boldsymbol{R}},\hat{\boldsymbol{t}})$. $\hat{\mathcal{I}}$ can then be visually matched with the previously obtained image segmentation $\mathcal{I}^\text{sem}$ by calculating a distance metric between both images and interpreting the result as a loss value. By adjusting $\hat{\boldsymbol{R}}$ and $\hat{\boldsymbol{t}}$ to minimize the visual difference between these images, and therefore minimizing the loss value, we actively determine the optimal camera parameters. An advantage of the Nelder-Mead method lies in its use of a simplex to define the initial search space, which then moves and shrinks towards the function minimum. By scaling the initial simplex bounds according to the expected measurement variance of the initial guess, we can easily define a sensible search space covered by the starting simplex. Note that other gradient-free approaches may be used as well.
\paragraph{Initialization}
In order to start the optimization, the Nelder-Mead method requires an initial parameter set. The convergence speed of the algorithm and its ability to find an optimal solution depend on the initial parameter set and the search space.
Therefore, an initial guess for the camera position and orientation is required.
We furthermore constrain the search space of the pitch and yaw angle to $\pm 5^\circ$, while we set the roll angle to zero. We expect our method to also work for non-vanishing roll angles, but in our use cases, no camera roll is desired and can be achieved by proper mounting. This is a reasonable simplification since potential camera roll is often not desired for automotive applications.
~The yaw angle can be measured with a compass, while the pitch can be determined with a spirit level. The maximal position offsets are constrained to $\pm2.5\,\si{\meter}$, i.e., the GPS accuracy of a typical smartphone GPS receiver.
We argue that an initial guess with such accuracy can be easily obtained for most practical situations and is sufficient to obtain a good solution for the extrinsic calibration problem, as shown in Sections~\ref{sec:ep} and~\ref{sec:rd}.
However, it is possible to extend the search space at a tolerable increase in calculation effort by combining the Nelder-Mead method with a coarse grid search to estimate parameters that were not measured and are therefore not known. If an IV sensor-to-sensor calibration is desired, no initial measurements are required in most cases, as the convergence range of our approach is sufficient to bridge the translation and rotation between both sensors. Therefore, they can be set to be zero vectors in such cases. After setting the initial guess, we also have to calculate the initial simplex, which will converge to the optimal solution. To this end, we generate 6 values (as a simplex of dimension $n$ is described by $n+1$ points) by adding and subtracting the maximal constrained parameter range for each pair of simplex vertices and for each parameter. 
\paragraph{Optimization Loop}
The Nelder-Mead method is invoked until the loss distance between two steps falls below $10^{-4}$ in order to estimate $\hat{\boldsymbol{R}}$ and $\hat{\boldsymbol{t}}$.
The loss target we optimize is defined as
\begin{equation}
	\label{eq:Loss}
	\mathcal{L}(\hat{\mathcal{I}}, \mathcal{I}^\text{sem}) = \beta \sum_{u=0}^{h-1} \sum_{v=0}^{w-1} \mathcal{V}_{u,v}\, \ell_{\mathrm{L2}}(\hat{\mathcal{I}}_{u,v},\,\mathcal{I}^\text{sem}_{u,v})
\end{equation}
with the L2 loss term $\ell$
\begin{equation}
    \ell_{\mathrm{L2}}(\boldsymbol{a},\, \boldsymbol{b})=(\boldsymbol{a}- \boldsymbol{b})^T(\boldsymbol{a}-\boldsymbol{b}).
\end{equation}
Here, $\beta$ is a normalization factor, $w$ and $h$ denote the width and height of the images,
~and $\mathcal{V}_{u,v}$ is a function checking the validity of the classes. If one of the images contains an invalid class for a pixel, $\mathcal{V}_{u,v}=0$, i.e., the pixel is ignored for loss calculations. Otherwise, $\mathcal{V}_{u,v}=1$ implies a contribution to the final loss. It should be noted that, while in theory, any distance metric may be used for $\ell$, we decided to use an L2 loss, as this loss is commonly used as an image similarity measure. Finally, the optimization performed by the Nelder-Mead method can be represented as
\begin{equation}
	\label{eq:OptimizationProblem}
	\hat{\boldsymbol{R}}_{\mathrm{opt}}, \hat{\boldsymbol{t}}_{\mathrm{opt}} = \arg \min_{\hat{\boldsymbol{R}}, \hat{\boldsymbol{t}}}\left\{ \mathcal{L}(\hat{\mathcal{I}}, \mathcal{I}^\text{sem}) \right\} .
\end{equation}
\paragraph{Appearance Matching and Masking}
Before optimizing the camera pose, we want to closely match the general appearance of the rendered image $\hat{\mathcal{I}}$ to the target $\mathcal{I}^\text{sem}$ as well as reduce rendering artifacts. For that reason, we  first aim to minimize the appearance of background pixels due to the sparsity of points in the 3D model. This is achieved by calculating the distance $d_i$ of every point in $\mathcal{M}^{3\text{D}}$ to the camera position. Then, each point in the 3D model is rendered as a circle of radius $r_i=\lambda/d_i$, where $\lambda$ is a scaling factor that depends on the point cloud density and can be determined empirically by rendering a test view and increasing $\lambda$ until the appearance of the rendered view roughly matches the target segmentation image. \ReviewChange{This adaptive point size approach also adds the benefit of limiting the impact of varying line counts between different lidar models, as lidars with fewer lines can still be densely rendered by increasing $\lambda$}. Secondly, pixels, which still cannot be assigned with a semantic label corresponding to a static object, e.g., due to point cloud sparsity, finally are classified as invalid so that these pixels do not contribute to the loss when estimating $\hat{\boldsymbol{R}}$ and $\hat{\boldsymbol{t}}$. This is achieved by storing this information in the mask elements $\mathcal{V}_{u,v}$ and masking the loss. In order to compensate for the potentially different amount of pixels contributing to the loss between two optimization steps, a normalization factor $\beta=\mathrm{numel}(\mathcal{V})/\sum_{u, v}\mathcal{V}_{u,v}$ is introduced, where $\mathrm{numel}$ denotes the element count of an array. Without $\beta$, a change of the camera pose could potentially reveal more surface area of a masked dynamic object. This may result in a lower loss due to fewer non-masked pixels contributing to it. At the same time, the pose may have changed away from the true pose, which should imply a higher loss, thus creating a need for loss normalization.
\paragraph{Result Verification}
\label{lbl:verification}
After performing an optimization step, we obtain the initial calibration result. However, as the loss function is highly non-convex with respect to the calibration parameters, this result may represent a local minimum. To avoid local minima, the optimization is restarted twice using the previous result  as the new start value while also lowering the convergence threshold from $10^{-4}$ to $10^{-6}$ during the last optimization step. The parameter set resulting in the lowest loss is chosen.

To check that the resulting parameters are indeed optimal, a small additive noise is added to the initial guess in order to exit a potentially found local minimum of the loss. The scale of the additive noise can be based on the measurement accuracy of the initially measured camera pose. Repeating the optimization with noisy initial values can then be used to discard sub-optimal parameter sets with high final loss values, which are caused by the non-convexity of the loss function.

\section{EVALUATION}

We evaluate our approach on infrastructure scenarios by observing a virtual intersection environment using the CARLA simulator~\cite{Dosovitskiy17}, as well as on a real-world environment based on the KITTI Odometry benchmark~\cite{geiger2013vision}. Next, we describe the evaluation protocol including the scene processing used for benchmarking and finally present our results. 

\subsection{Evaluation Scenes}

\begin{figure}[!t]
    \centering
    \subfloat[Ideal segmentation]{
         \includegraphics[trim=0cm 2.5cm 0cm 0.75cm, clip, width=0.8\linewidth]{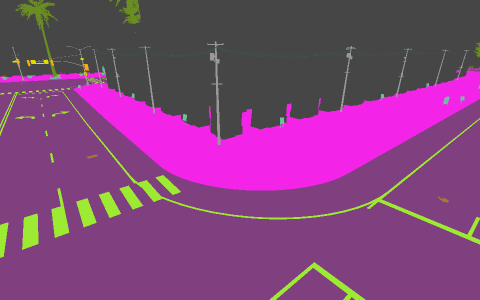}
         \label{fig:carla1}}
    \hfil
    \subfloat[Rendered point cloud view]{
         \includegraphics[trim=0cm 2.5cm 0cm 0.75cm, clip, width=0.8\linewidth]{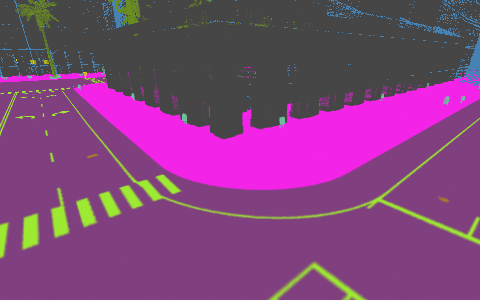}
         \label{fig:carla2}}
    \caption{CARLA simulator \cite{Dosovitskiy17} sample result (Town10HD). (a) depicts an ideal semantic image segmentation of a camera sensor. (b) shows a rendered view of a semantically segmented point cloud.}
    \label{fig:carla}
\end{figure}
Similar to related work like \cite{datondji2016rotation} and \cite{ataer2014calibration}, we use predefined scenes for evaluation. For both CARLA and KITTI, two distinct scenes were defined. For CARLA, we configured urban intersection views from an elevated static perspective to mimic real use cases for roadside infrastructure. Meanwhile, for KITTI, one time frame of two different urban driving sequences was chosen, as we only need one time frame at which the camera will be calibrated. In this case, however, we merged 250 lidar scans around the timestamp of the chosen frame to reconstruct the 3D environment. In the case of CARLA, the segmented point clouds, as well as the segmented camera images of the defined scenes, were generated using the built-in maps. Given completely correct segmentation maps, it is easier to assess the performance of our approach. \ReviewChange{The virtual lidar was configured to have 128 lines and a vertical field of view of $90^\circ$ with the center of the FoV aligned to the ground plane. The virtual segmentation camera used a resolution of 960x600 with a horizontal FoV of $90^\circ$}. For the KITTI data, the Cylinder3D~\cite{zhu2021cylindrical} model, trained with Semantic-KITTI labels \cite{behley2019semantickitti}, was used to generate the point cloud segmentation. Therefore, no ground truth data available in Semantic-KITTI was used. Moreover, the image segmentation was obtained with OCRNet~\cite{yuan2021segmentation}, trained with Cityscapes labels \cite{cordts2016cityscapes}. \ReviewChange{The sensor setup used for recording the raw data passed to the network is described in the KITTI paper~\cite{geiger2013vision}}. As the image segmentation network provided sub-optimal results in some cases, we manually corrected the segmentation to additionally evaluate for a \textit{semi-automatic} setting. In our \textit{automatic} setting, the original image segmentation was utilized without any modification. In all KITTI scenes, road and sidewalk labels were merged in both domains to circumvent noisy borders between ground points which would hinder the optimization. At the same time, the dynamic object removal mentioned in Sec.\ref{lbl:pclseg} was limited to the driveable area covered by the measurement vehicle, which helped retain parked vehicles in the point cloud. This helped to create more visual comparison areas not masked out by $\mathcal{V}_{u,v}$.

\begin{figure}[!t]
    \centering
    \subfloat[Rectified camera image]{
         \includegraphics[width=0.8\linewidth]{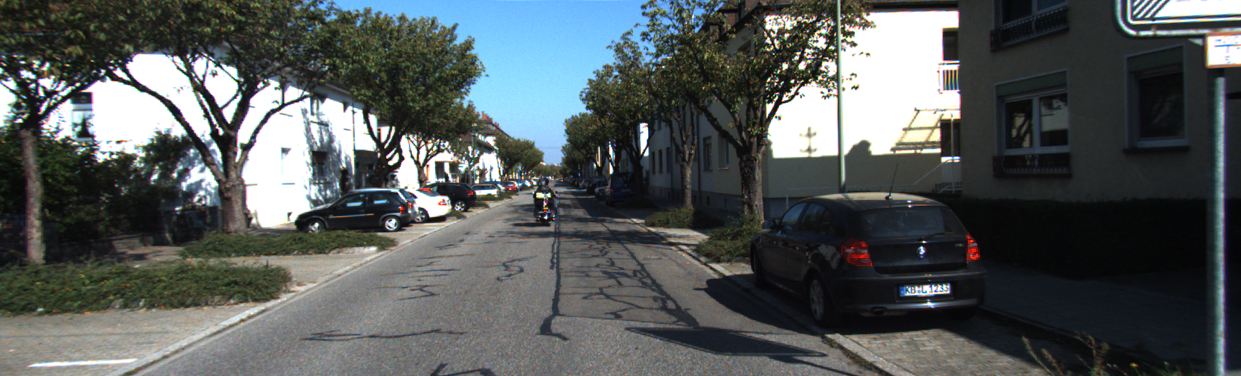}
         \label{fig:kitti1}}
    \hfil
    \subfloat[Image segmentation]{
         \includegraphics[width=0.8\linewidth]{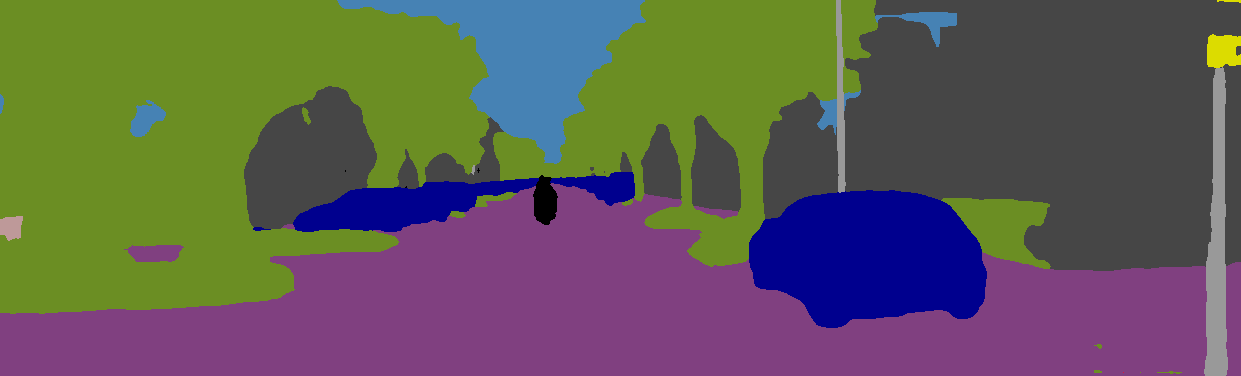}
         \label{fig:kitti2}}
    \hfil
    \subfloat[Rendered point cloud view]{
         \includegraphics[width=0.8\linewidth]{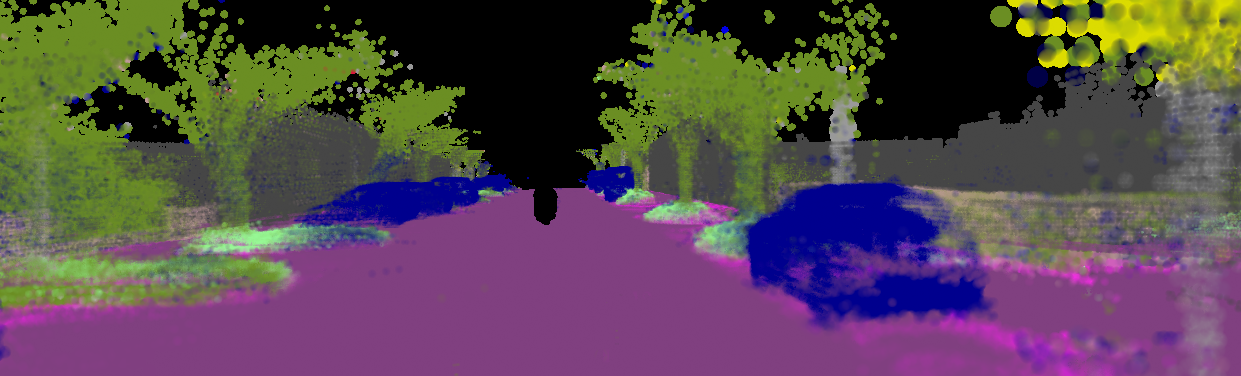}
         \label{fig:kitti3}}
    \caption{KITTI \cite{geiger2013vision} dataset result (Sequence 00 fully automated). We can see a noticeable visual difference between the segmentation image and the rendered point cloud, which is caused by point cloud sparsity as well as incorrect labels. However, they still share enough common labelfeatures to be reliably matched, as shown in Table \ref{tbl:mce}.}
    \label{fig:kitti}
\end{figure}

Furthermore, during CARLA evaluation, the point cloud view was rendered with a sky background, as visible in Fig. \ref{fig:carla2}. In order to only match the sky region common between both images, sky regions in the rendered view not matching sky regions in the target were masked out, as these pixels are caused by point cloud sparsity. Another efficient measure consisted of only using the bottom half of the image for registration, as in the KITTI lidar data, the point cloud was cropped in the height dimension, resulting in a trimmed view of the scene which introduced high discrepancies in the upper half of the image. This can be clearly seen in Fig. \ref{fig:kitti3}. Finally, the point cloud was cropped to a 75m radius around the initial location to reduce the memory footprint, in the case of KITTI evaluation. The lidar segmentation view was rendered with Pytorch3D \cite{ravi2020pytorch3d}.

\subsection{Evaluation Protocol}
\label{sec:ep}
As the ground truth calibration for both evaluations is known, a random initialization procedure was performed to assess the convergence behavior of the optimization approach. This was implemented by initializing the optimization algorithm with noisy ground truth, where the additive noise was uniformly sampled from the interval $[-2.5\,\si{\meter};2.5\,\si{\meter}]$ for positional offsets and $[-5^\circ;5^\circ]$ for rotational offsets, as mentioned in Sec. \ref{lbl:cdreg}. By using this approach, we can also verify the usability of our approach, as such offsets could appear in field measurements. After the completion of the optimization, the remaining offsets were interpreted as the calibration quality. As all coordinate systems used are scaled to meters, the positional offsets are measured in centimeters. The algorithm is invoked 30 times on each scene and the 10 parameter sets resulting in the lowest loss $\mathcal{L}(\hat{\mathcal{I}}, \mathcal{I}^\text{sem})$ are considered as acceptable results to filter out results that would be considered outliers
, as described in the Result Verification of Sec. \ref{lbl:cdreg}.\\
 We then report the mean vector norm of the displacement error vector $\overline{\lVert\Delta_{\boldsymbol{t}}\rVert} = \overline{\lVert\hat{\boldsymbol{t}_\mathrm{ext}} - \boldsymbol{t}_\mathrm{ext}\rVert}$ as well as the mean absolute angular error $\overline{|\Delta_{\alpha}|}$. Furthermore, to evaluate our choice of the L2 loss as a distance metric, we also repeated the previously described evaluation protocol with the Huber loss \cite{huber1992robust}
 \begin{align}
	\ell_{\mathrm{Huber}}(\boldsymbol{a},\, \boldsymbol{b}) = \begin{cases}
	\frac{1}{2}(\boldsymbol{a}- \boldsymbol{b})^T(\boldsymbol{a}-\boldsymbol{b}),&\lVert\boldsymbol{a}-\boldsymbol{b}\rVert<\delta\\
	\delta(\lVert\boldsymbol{a}-\boldsymbol{b}\rVert - \frac{\delta}{2}),&\mathrm{otherwise}
	\end{cases}
\end{align}
 For results using the Huber loss, we set the loss threshold $\delta$ to 0.3.

\begin{table}[h]
    \caption{Evaluation results.}
    \label{tbl:mce}
    \centering
    \begin{tabular}{lcccc}
        \toprule
        \textbf{Loss} & \multicolumn{2}{c}{L2} & \multicolumn{2}{c}{Huber}\\
        \midrule
        \textbf{Metric} & $\overline{\lVert\Delta_{\boldsymbol{t}}\rVert}$ & $\overline{|\Delta_{\alpha}|}$ & $\overline{\lVert\Delta_{\boldsymbol{t}}\rVert}$ & $\overline{|\Delta_{\alpha}|}$\\
        \midrule
        \multicolumn{5}{l}{\textbf{Fully automated CARLA-simulated infrastructure}}\\
        Town10HD& $1.5\,\si{\cm}$&$0.03^\circ$ & $1.5\,\si{\cm}$&$0.02^\circ$\\
        Town01& $6.0\,\si{\cm}$&$0.11^\circ$ & $6.2\,\si{\cm}$&$0.10^\circ$\\
        \midrule
        \multicolumn{5}{l}{\textbf{Semi-automated KITTI}}\\
        Sequence 00 & $12.7\,\si{\cm}$&$0.28^\circ$ & $15.3\,\si{\cm}$&$0.26^\circ$\\
        Sequence 09 & $20.6\,\si{\cm}$&$0.40^\circ$ & $19.3\,\si{\cm}$&$0.42^\circ$\\
        \midrule
        \multicolumn{5}{l}{\textbf{Fully automated KITTI}}\\
        Sequence 00 & $20.2\,\si{\cm}$&$0.34^\circ$ & $20.3\,\si{\cm}$&$0.33^\circ$\\
        Sequence 09 & $18.6\,\si{\cm}$&$0.40^\circ$ & $19.6\,\si{\cm}$&$0.43^\circ$\\
        \bottomrule
    \end{tabular}
\end{table}

\subsection{Results Discussion}
\label{sec:rd}

Table~\ref{tbl:mce} summarizes our results for both evaluations and examined loss functions. 
For the CARLA evaluation, our translation error is below 7\,\si{\cm} for both scenes, while the rotation error is at most $0.11^\circ$.
~We observed that the errors mostly stem from point cloud sparsity, as can be seen in Fig. \ref{fig:carla}. It should also be noted that, due to the rasterized representation of an image, we cannot distinguish image translations below a single pixel, therefore very small changes of under 1\si{\cm} in camera translation and \ang{0.01} in camera rotation may not be visible in the rendered view. The real-world KITTI scenes also perform very well. While both the point cloud labels and image segmentation contain incorrect labels and, in the case of point clouds, incorrectly measured data points, the final calibration quality is still very high, especially if used for infrastructure sensor calibration. Furthermore, we tested fully automatically labeled KITTI scenes by not manually correcting the image segmentation.
~The performance in these scenes dropped only slightly or even stayed in the same range and is therefore comparable with the semi-automatic use case. However, it should be noted that this performance highly depends on the quality of the sensor data and the dataset used for training the segmentation network. Comparing both loss types, we can see only minor differences, which can be explained by the random initialization and testing approach. Thus, both losses are equally suitable for the task.
\subsection{Comparison with Related Approaches}

As our approach is quite different from previous works in this area and no common evaluation protocols are defined, it is difficult to directly compare performance metrics, so we only look at qualitative results. Considering Horn et al. \cite{horn2021online}, who achieve translation errors of $20.8\,\si{\cm}$ and rotation errors of $0.26^\circ$ on the KITTI dataset, but require sensor movement, we get equally good results (see Table \ref{tbl:mce}) without sensor movement. The authors of the mobile stereo camera SLAM approach in \cite{ataer2014calibration} report translation errors in the range of $6\si{\cm}-67 \si{\cm}$ and rotation errors between $1.5^\circ$ and $5.4^\circ$ on their own data. Comparably, our positional errors are on the lower end of the spectrum, while our rotation errors are smaller by a factor of 10. We, therefore, conclude that our approach is competitive with other methods and can be used for infrastructure and IV sensor calibration.

\section{CONCLUSION}

We presented an extrinsic camera calibration approach for both infrastructure and intelligent vehicle cameras.
~In our approach, we obtained a semantic segmentation of both IV-collected lidar data as well as camera data. This segmented data was then used to find an optimal calibration by using an optimization loop to match the appearance of the image segmentation with a rendered view of the segmented lidar data, effectively performing cross-domain registration. The algorithm has been evaluated on simulated CARLA \cite{Dosovitskiy17} scenes and real sensor data from the KITTI dataset \cite{geiger2013vision} to show its feasibility in a practical application, showing equally good or better results than existing approaches, which, however, are more constrained in their application.

\addtolength{\textheight}{-9cm}   



%
%
%


\bibliographystyle{ieeetran}
\bibliography{IEEEexample}

\end{document}